\documentclass[5p,times]{elsarticle}

\usepackage{amsmath,amssymb} 
\usepackage{amsfonts}        

\usepackage{graphicx}         
\usepackage{xcolor}           
\usepackage{caption}
\usepackage{array}            
\usepackage{booktabs}         
\usepackage{multirow}         
\usepackage{tabularx}         
\usepackage{caption}          
\usepackage{subcaption}       

\usepackage{algorithm}        
\usepackage{algpseudocode}    

\usepackage{float}            
\usepackage{enumitem}         
\usepackage{url}              
\usepackage[bookmarks=false, colorlinks=true, linkcolor=blue, citecolor=blue, urlcolor=blue]{hyperref}
\UseRawInputEncoding
\begin{document}
\begin{frontmatter}


\title{In-Situ Melt Pool Characterization via Thermal Imaging for Defect Detection in Directed Energy Deposition Using Vision Transformers}


\author[a]{Israt Zarin Era} 
\author[a]{Fan Zhou}
\author[a]{Ahmed Shoyeb Raihan}
\author[a]{Imtiaz Ahmed}
\author[c]{Alan Abul-Haj}
\author[d]{James Craig}
\author[a,b]{Srinjoy Das}\corref{cor2}
\ead{srinjoy.das@mail.wvu.edu }
\author[a]{Zhichao Liu\corref{cor1}}
\ead{zhichao.liu@mail.wvu.edu }

\address[a]{Industrial \& Management Systems Engineering, Address, West Virginia University, Morgantown 26506, USA}
\address[b]{School of Mathematical and Data Sciences, West Virginia University, Morgantown, WV 26506, USA}
\address[c]{ARA Engineering, Inc, Sedona, AZ 86336, USA}
\address[d]{Stratonics, Inc, Lake Forest, CA 92630, USA}

\begin{abstract}
Directed Energy Deposition (DED) has significant potential for rapidly manufacturing complex and multi-material parts. However, it is prone to internal defects, such as lack of fusion porosity and cracks, that may compromise the mechanical and microstructural properties, thereby, impacting the overall performance and reliability of manufactured components. This study focuses on in-situ monitoring and characterization of melt pools closely associated with internal defects like porosity, aiming to enhance defect detection and quality control in DED-printed parts. Traditional machine learning (ML) approaches for defect identification require extensive labeled datasets. However, in real-life manufacturing settings, labeling such large datasets accurately is often challenging and expensive, leading to a scarcity of labeled datasets. To overcome this challenge, our framework utilizes self-supervised learning using large amounts of unlabeled melt pool data on a state-of-the-art Vision Transformer-based Masked Autoencoder (MAE), yielding highly representative embeddings. The fine-tuned model is subsequently leveraged through transfer learning to train classifiers on a limited labeled dataset, effectively identifying melt pool anomalies associated with porosity. In this study, we employ two different classifiers to comprehensively compare and evaluate the effectiveness of our combined framework with the self-supervised model in melt pool characterization. The first classifier model is a Vision Transformer (ViT) classifier using the fine-tuned MAE Encoder's parameters, while the second model utilizes the fine-tuned MAE Encoder to leverage its learned spatial features, combined with an MLP classifier head to perform the classification task. Our approach achieves overall accuracy ranging from 95.44\% to 99.17\% and an average F1 score exceeding 80\%, with the ViT Classifier outperforming the MAE Encoder Classifier only by a small margin. This demonstrates the potential of our framework as a scalable and cost-effective solution for automated quality control in DED, effectively utilizing minimal labeled data to achieve accurate defect detection.

\end{abstract}

\cortext[cor1]{Corresponding authors.}

\end{frontmatter}

\section{Introduction}\label{intro}
\begin{figure}[t]\vspace*{4pt}
    \centering
    \includegraphics[width=\columnwidth]{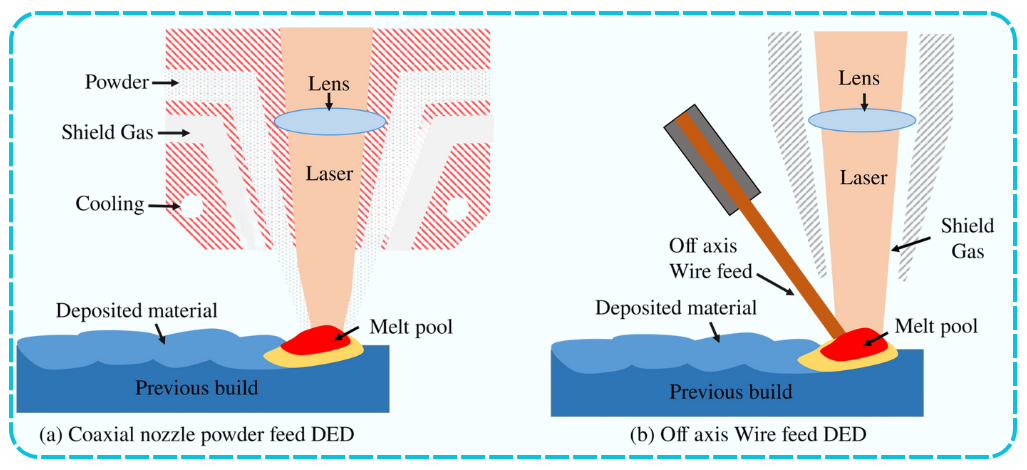}
    \caption{Cross-sectional view of Directed Energy Deposition (DED) processes redrawn from \cite{era2023machine}: (a)Powder-feed and (b)Wire-feed.}
    \label{fig1.1}
\end{figure}

Directed Energy Deposition (DED) \cite{ahn2021directed} is widely used in aerospace and biomedical industries for its ability to precisely repair and build high-value components. This laser-based additive manufacturing process utilizes a focused laser beam to melt and fuse metal powder or wire layer by layer, creating a moving melt pool \cite{thompson2015overview} at the laser-material interface as shown in Figure \ref{fig1.1}. However, these intricate interactions can sometimes introduce defects like porosity, which compromises the mechanical integrity and reliability of the final parts \cite{svetlizky2021directed}. Porosity in DED components arises primarily from entrapped gas bubbles within the melt pool, originating either from gases trapped in the feedstock or from incomplete fusion between layers \cite{zhang2024pore}. As the melt pool solidifies, these bubbles may become trapped, forming voids that impact the material's strength. The melt pool’s characteristics such as size, shape, and temperature distribution are crucial in shaping the microstructural and mechanical properties of the part \cite{svetlizky2021directed}. In-situ characterization of the melt pool provides valuable insights into defect formation, allowing manufacturers to identify and mitigate porosity and other defects \cite{tang2020review}. By monitoring melt pool dynamics during the DED process, manufacturers can also enhance quality control and ensure the reliability of their final products.

In recent years, significant efforts have been made to develop in-situ monitoring and characterization methods for melt pool dynamics in laser additive manufacturing (LAM), especially for identifying internal defects within directed energy deposition (DED). \cite{khanzadeh2018porosity} introduced a real-time porosity detection framework using Self-Organizing Map (SOM) clustering on thermal melt pool images, achieving 96\% accuracy. Similarly, \cite{tian2021deep} proposed a fusion approach combining PyroNet (CNN) and IRNet (LRCN) for porosity prediction via decision-level integration of pyrometer and infrared images. \cite{khanzadeh2019situ} reached 98.44\% accuracy using thermal history and morphological features with a K-Nearest Neighbor (KNN) model. Meanwhile, \cite{zhao2021automated} employed a variational autoencoder (VAE) with Gaussian mixture modeling (GMM) and K-means clustering to detect melt pool anomalies in DED, reporting 94.52\% accuracy. In laser wire DED, \cite{asadi2024process} leveraged YOLO-based CNNs for real-time segmentation and analysis, improving monitoring precision. \cite{ertay2021process} used a high-dynamic range camera with a KNN classifier to classify stability zones, achieving a 13\% error rate. In the mean time, multi-sensor fusion and data-driven models have also shown promise. \cite{gaikwad2022multi} integrated multiple sensors with machine learning models like SVM, KNN, Random Forest, and MLP, achieving a 90\% true positive rate while \cite{ouidadi2023real} applied an unsupervised online learning method in laser metal deposition (LMD), reaching 76\% accuracy with K-means and 97\% with SOM for classifying images as healthy" or anomalous." Another work by \cite{kong2023development} introduced a defect diagnosis system in DED using a degree of irregularity (DOI) feature index, achieving up to 96.92\% accuracy for balling defects, outperforming STFT-based CNN models.
More recent developments continue to push the boundaries of in-situ monitoring accuracy. \cite{zeng2024classification} proposed FixConvNeXt, achieving 99.1\% accuracy in real-time melt pool state identification for L-DED while reducing computational costs. \cite{chen2023situ} demonstrated an acoustic-based approach using CNNs for defect detection in DED, reaching 89\% overall accuracy and 98\% AUC-ROC for keyhole pore detection. \cite{bappy2022morphological} introduced a Gaussian SVM classifier for layer-wise anomaly detection in DED, achieving up to 96.38\% accuracy and a 95.34\% F-score, surpassing benchmarks in efficiency. \cite{yuan2022method} utilized DenseNet-39 with high-speed camera images, achieving 99.3\% accuracy in classifying melt pool states as stable or unstable, reducing computational demands. \cite{abranovic2024melt} employed a ConvLSTM autoencoder with high-speed camera video data for detecting defects like wire dripping and arcing, effectively enabling real-time anomaly detection in DED. We summerize the different learning methodologies and models employed for defect detection tasks through in-situ melt pool data within the DED process in Table \ref{tab:literature_review}.

\captionsetup{skip=2pt}
\begin{table*}[h!]
\centering
\caption{Summary of literature on In-Situ Monitoring and Defect Detection via Melt Pool data in DED.}
\begin{tabularx}{\textwidth}{lllc}
\hline
\textbf{Tasks} &  \textbf{Method} & \textbf{Models} & \textbf{Study} \\
\hline
Real-time porosity detection  & Unsupervised & Self-Organizing Map (SOM) & \cite{khanzadeh2018porosity} \\

Porosity prediction through multi-sensor data fusion  & Supervised & PyroNet (CNN), IRNet (LRCN) & \cite{tian2021deep} \\

Classification of melt pool from thermal \& morphological features  & Supervised & K-Nearest Neighbor (KNN) & \cite{khanzadeh2019situ} \\

Melt pool anomaly detection  & Unsupervised & VAE,GMM, K-means & \cite{zhao2021automated} \\

Real-time segmentation \& classification of melt pools  & Supervised & YOLOv5, YOLOv8 & \cite{asadi2024process} \\

Classification of melt pool stability  & Supervised & K-Nearest Neighbor (KNN) & \cite{ertay2021process} \\

Porosity detection from multi-sensor data fusion & Supervised & SVM, KNN, MLP, RF & \cite{gaikwad2022multi} \\

Real-time defect detection from melt pool & Unsupervised & K-means, SOM & \cite{ouidadi2023real} \\

Defect detection from melt pool irregularities & Supervised &SIFT-based 2D CNN & \cite{kong2023development} \\

Real-time melt pool state identification  & Supervised & FixConvNeXt & \cite{zeng2024classification} \\
Defect detection through acoustic signal  & Supervised & KNN, SVM, CNN & \cite{chen2023situ} \\
Layer-wise melt pool anomaly detection  & Supervised & SVM classifier & \cite{bappy2022morphological} \\

Melt pool state classification  & Supervised & DenseNet-39 & \cite{yuan2022method} \\

Defect detection by melt pools  & Self-Supervised & ConvLSTM Autoencoder & \cite{abranovic2024melt} \\
\hline
\end{tabularx}
\label{tab:literature_review}
\end{table*}

Moreover, Table \ref{tab:literature_review} highlights that Convolutional Neural Network (CNN) based models \cite{abranovic2024melt, chen2024situ, zeng2024classification, tian2021deep, zhang2024machine, zhao2021automated, yuan2022method, asadi2024process} have been at the forefront of advancements in real-time melt pool monitoring and defect detection within DED process. While CNNs, often applied through supervised learning \cite{tian2021deep} in this domain, have shown success in defect detection using melt-pool features, they come with limitations. These models struggle to perform optimally when faced with data domains characterized by linearly inseparable features or fuzzy object boundaries \cite{mauricio2023comparing}. Additionally, acquiring large labeled datasets for melt pool monitoring is costly and time-intensive, which limits CNNs' effectiveness and often leads to overfitting. Transfer learning has \cite{zhuang2020comprehensive} emerged as a solution to this problem by leveraging pre-trained models on similar datasets to improve model performance on the smaller, domain-specific data. However, transfer learning is not always effective, especially when there is a significant domain shift or the data distribution differs substantially from the pre-trained model's original training set. Consequently, there is a need for more adaptable and versatile approaches that can utilize the abundance of unlabeled data generated in manufacturing processes.

Given these limitations of supervised learning models such as CNNs, self-supervised learning (SSL) \cite{misra2020self} has become an increasingly popular approach that reduces the reliance on large labeled datasets by enabling models to learn directly from the data through reconstruction. SSL trains models on pretext tasks that generate predictive signals from the data itself, allowing models to learn rich feature representations. These learned features can later be fine-tuned with smaller labeled datasets, making SSL a resource-efficient method in manufacturing contexts where labeled data is scarce. This capability is particularly valuable in Directed Energy Deposition (DED) processes, where labeled data is often limited, and traditional supervised models may struggle to generalize effectively on smaller datasets. 

Researchers have explored various SSL techniques, combining them with supervised classifiers or unsupervised clustering of latent features for defect identification tasks \cite{zhao2021automated, abranovic2024melt}. However, in the domain of melt pool thermal images for defect detection, unsupervised clustering may be less suitable since the latent features are not usually linearly separable, which can limit the effectiveness of identification without labeled data. Additionally, to apply a supervised classifier on latent features, it is essential to ensure high-quality feature generation to support accurate supervised classification. 
To overcome these constraints, we utilize Vision Transformers (ViTs) \cite{dosovitskiy2020image}, which are transforming the computer vision landscape with their superior capability to capture intricate spatial relationships and global dependencies within high-resolution images. Unlike CNNs, Vision Transformers (ViTs) divide images into patches and process them sequentially, allowing for enhanced feature extraction at both local and global levels. This capability is especially advantageous in applications requiring high detail and precision, such as melt pool analysis for defect detection. This study employs a state-of-the-art ViT-based Masked Autoencoder (MAE) \cite{he2022masked} in a self-supervised learning framework. The MAE model has formed the backbone of several recent foundation models \cite{kirillov2023segment, tong2022videomae} in computer vision. This model is used to learn spatial features directly from the unlabeled data. Inspired by the Large Language Models (LLM) \cite{devlin2018bert}, the MAE reconstructs missing parts from the masked portions of the input during training, thereby learning robust, representative features. These learned features are then leveraged to train classifiers on a small labeled dataset, enabling accurate identification of melt pool anomalies correlated with porosity. 
The summary of our contribution is: 
\begin{itemize}
    \item We develop a self-supervised learning approach combined with supervised classification which is well-suited to real-world LAM environments, where unlabeled data, such as thermal images of melt pools, is abundant, while labeled data remains scarce due to the intensive nature of physical experimentation required for labeling. 
    \item We minimize computational overhead by using a pre-trained ViT-based Masked Autoencoder (MAE); fine-tuning it on a substantial number of unlabeled thermal images captured from in-situ melt pool monitoring during DED printing.
    \item Our proposed framework employs Transfer Learning to utilize the fine-tuned model's parameters in training various classifiers on limited labeled data for accurate melt pool classification.
    \item To evaluate our approach, we compare two supervised classifiers: a standard ViT Classifier and a fine-tuned MAE Encoder paired with an MLP classifier, both trained on labeled data and utilizing the fine-tuned parameters and learned features respectively from the self-supervised MAE model. This comparison validates the effectiveness of our framework combining self-supervised learning with different supervised classification strategies by achieving reliable identification of melt pool classes.
\end{itemize}

The rest of the paper is organized into the following sections. In Section \ref{exp} we briefly describe the experimental setup and in-situ melt pool data collection used in this study. Section \ref{method} explains the methodology used in our approach with the self-supervised MAE and ViT Classifier models. In Section \ref{result} we demonstrate the obtained melt pool classification results from using both the classifiers with our setup. Finally, in Section \ref{conclu} we summarize our findings and contributions with final remarks. 

\section{Experimentation \& Data collection}\label{exp}
\begin{figure}[t]\vspace*{4pt}
    \centering
    \includegraphics[width=\columnwidth]{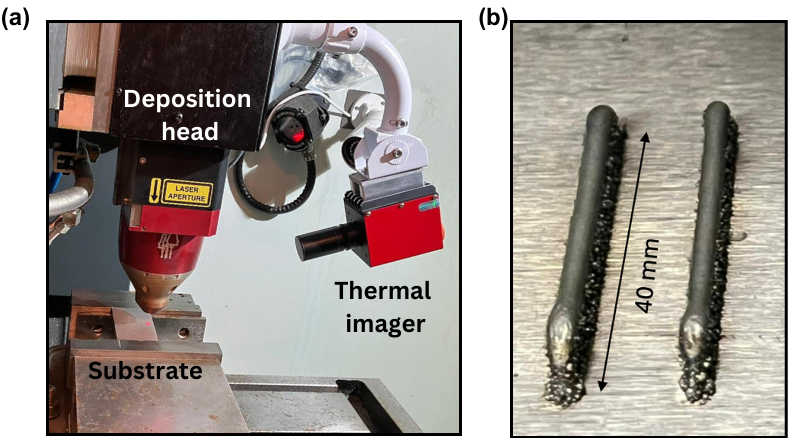}
    \caption{(a) The pyrometer setup inside the DED printing chamber; (b) Printed Samples.}
    \label{fig2.1}
\end{figure}

\begin{figure}[t]\vspace*{4pt}
    \centering
    \includegraphics[width=\columnwidth]{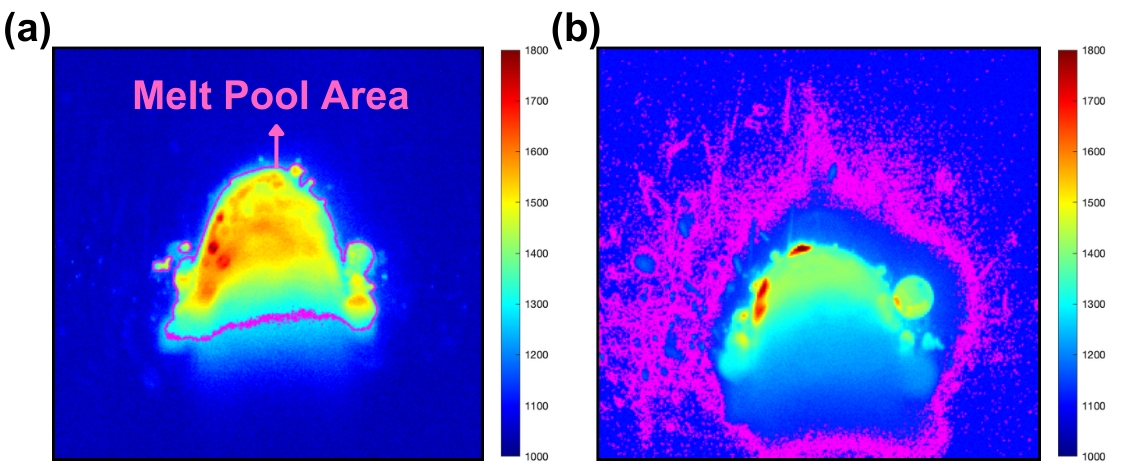}
    \caption{Examples of (a) A normal melt pool; (b) An abnormal melt pool.}
    \label{fig2.2}
\end{figure}

For data collection, we conduct the experiments using a customized powder-based DED system (AMBIT\textsuperscript{\texttrademark} Core DED, Hybrid Manufacturing Technology, TX, USA). We mount a thermal imager Pyrometer camera (ThermaViz\textsuperscript{\textregistered} TV200, Stratonics) inside the printing chamber for in-situ monitoring and melt pool data capture during printing. Since our DED setup is hybrid, including both additive and CNC machines, we do an off-axis installation of the camera at a viewing angle of 57$^\circ$ (as shown in Figure \ref{fig2.1}) to maintain a clear line of sight to the melt pool while avoiding obstruction from the laser head and deposition tool and ensuring the required working distance of $245 \pm 30$ mm. The focal length is fixed at 42 mm. We perform a positional and temperature calibration of the pyrometer camera before each experiment. A Neutral density filter is also calibrated and used to prevent pixel saturation during the in-situ recording by the pyrometer. In this study, we print two single-track thin-wall Inconel 718 samples, each measuring 40 mm in length and consisting of 10 layers in Figure \ref{fig2.1}, with the scanning direction reversed on every layer to create a linear reciprocating scan pattern. The variations in printing strategies and process parameters are summarized in Table \ref{tab1} below. Note that, we have fixed the exposure time of the pyrometer at 0.3 ms optimized by experimentation for both sample printing. The camera records the distribution of thermal gradient of the traveling melt pool during the printing and the frames can be retrieved as individual thermal images. The melt pool zone is defined as the region where the temperature exceeds the material's melting point. For Inconel 718, this temperature threshold is $1336^\circ$ Celsius. The boundary of the melt pool surface is represented by an isotherm, as illustrated in Figure \ref{fig2.2}. Each pixel in the thermal images represents the corresponding temperature of the melt pool zone and the rest of the area within the field of view of the camera. The transmissivity of Inconel 718 is considered negligible as the material is highly opaque. The two-color pyrometer reduces emissivity effects by measuring infrared radiation at two wavelengths, with pre-calibration using a manufacturer-provided lamp and slope correction adjustments as needed.
Then, we perform a $360^\circ$ X-ray Computed Tomography (XCT) (SkyScanner1272) scan on Sample 2 (from Table \ref{tab1}) to identify the locations of the inter-layer pores and label the corresponding melt pool images captured at those specific locations. We find a total of 76 melt pools associated with internal pores and label them as class `1' and the rest 1371 normal melt pools as class `0'. Examples of normal and abnormal melt pool thermal images are shown in Figure: \ref{fig2.2}. 

\begin{table}[h!]
\caption{Experimental parameters and dataset characteristics.}
\centering
\renewcommand{\arraystretch}{1} 
\begin{tabularx}{\linewidth}{>{\centering\arraybackslash}X >{\centering\arraybackslash}X >{\centering\arraybackslash}X >{\centering\arraybackslash}X >{\centering\arraybackslash}X >{\centering\arraybackslash}X}
\hline
\textbf{Samples} & \textbf{Laser power (W)} & \textbf{Scanning speed (mm/s)} & \textbf{Frame rate (Hz)} & \textbf{Number of images} & \textbf{Labeled} \\
\hline
1 & 800 & 10 & 370 & 7812 & \ding{55} \\ 
2 & 700 & 10 & 64 & 1447 & \ding{51} \\ 
\hline
\end{tabularx}
\label{tab1}
\end{table}

\section{Methodology}\label{method}
Our framework can be divided into two stages. First, we fine-tune a pre-trained MAE model on unlabeled melt pool images in a self-supervised manner. Next, we apply these fine-tuned model parameters to train classifier models on the limited labeled data, enabling binary classification of the melt pool images. These steps are explained in detail in the following sections. 

\subsection{Masked AutoEncoder (MAE)}\label{mae}
The Masked Autoencoder (MAE) is a scalable self-supervised learning framework that learns meaningful spatial features from unlabeled image data through image reconstruction. Like a typical autoencoder, MAE maps input data to a latent representation and then reconstructs the original data from this representation. However, MAE differs from an autoencoder as it uses only partial information from the input during encoding, reconstructing the complete image based on this limited data. This innovative masking strategy effectively reduces computational overhead, one of the main challenges in vision models. The model employs an asymmetric architecture: the encoder processes only the visible patches, while a lightweight decoder reconstructs the full image using the latent representation of the visible patches and the mask tokens. The details are explained below.

\subsubsection{MAE Encoder}\label{mae_enc}
The encoder in MAE is based on a Vision Transformer (ViT) \cite{dosovitskiy2020image} later described in section \ref{vit_class} but without the classifier MLP head (Figure \ref{fig2.4}). Unlike traditional autoencoders, it operates only on the visible, unmasked patches of an image while the rest are masked. MAE divides each image into non-overlapping patches. A subset of these patches (covering only 25\% of the entire image) is randomly selected to remain visible. This sampling follows a uniform distribution to prevent any spatial bias, such as a concentration of visible patches near the image center. As the remaining 75\% of the images are not supplied to the encoder, the reconstruction task becomes more challenging and this cannot be done successfully by simply relying on neighboring pixels like CNNs or neighboring patches like a typical ViT. To address this each visible patch in the MAE is embedded using linear projection with positional embeddings, following which it is processed through transformer blocks.

\subsubsection{MAE Decoder}\label{mae_dec}

 The MAE decoder is also a ViT transformer, that takes the full set of tokens of the original images, which includes both the encoded visible patches and additional mask tokens that represent the missing patches to be predicted. Each mask token is a shared, learned vector indicating a missing part of the image. Positional embeddings are added to all tokens to ensure the decoder understands the spatial location of each mask token within the image. The decoder is used solely during pre-training/fine-tuning for the reconstruction task, while the encoder's learned features are used later for recognition tasks. 
 This asymmetry reduces the overall inference time, making the model more efficient for large-scale applications. At the end of the decoder, there is a linear projection layer that maps the decoder's output embeddings to pixel values. This layer translates the high-dimensional feature representations back into the pixel space of the masked patches. The model uses Mean Square Error (MSE) (equation \ref{eqmse}) as the reconstruction loss between the predicted pixel values and the actual pixel values of the original image for the masked patches. 
\vspace{-20pt}
\begin{equation}\label{eqmse}
\vspace{-20pt}
\text{MSE} = \frac{1}{N} \sum_{i=1}^{N} (\text{predicted}_i - \text{target}_i)^2
\end{equation}
Where, $N$ is the total number of samples, $ \text{predicted}_{\text{i}} $
 refers to the predicted value for the i-th sample and $\text{target}_{\text{i}} $refers to the target or actual value for the i-th sample.

The MAE pre-training process is designed to be efficient and avoids the need for specialized sparse operations. First, each input patch is converted into a token through linear projection, with positional embeddings added. The tokens are then randomly shuffled, and a portion is removed according to the masking ratio, resulting in a subset of tokens for the encoder. This process, equivalent to sampling patches without replacement, creates a compact input for the encoder. After encoding, mask tokens are added to the encoded patches, and the shuffled list is restored to its original order to align each token with its corresponding target. The decoder processes this complete set of tokens, with positional embeddings and reconstructs the target. This straightforward approach incurs minimal computational overhead, as the shuffling and unshuffling steps are computationally light.

\subsection{Self-supervised Framework with classifiers}\label{framework}
\begin{figure*}[t]\vspace*{4pt}
    \centering
    \includegraphics[width=\textwidth]{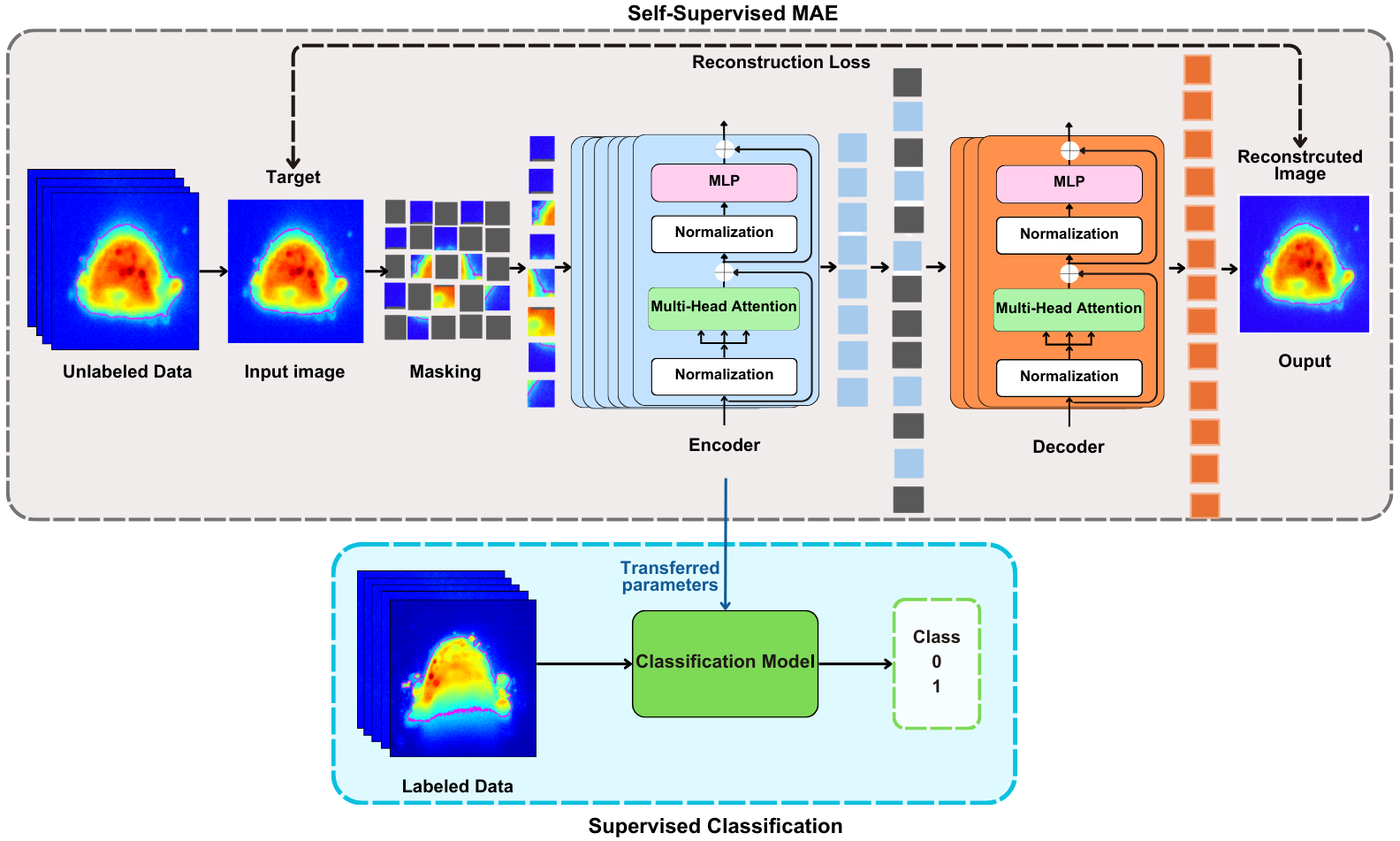}
    \caption{Overview of the framework for in-situ characterization of melt pool images using a self-supervised MAE with classifier models.}
    \label{fig2.3}
\end{figure*}

In this study, we exploit the self-supervised learning capabilities of MAE to capture valuable spatial information from unlabeled melt pool images. Instead of training the model from scratch, we leverage a pre-trained MAE base model that has already undergone extensive training on the ImageNet-1k dataset for 800 epochs using a high-performance setup \cite{he2022masked}. This pre-trained model has demonstrated superior performance over previous ViT-based models in image recognition tasks. Therefore, we fine-tune this pre-trained MAE with our unlabeled melt pool data (following the steps in Algorithm \ref{alg1}), significantly reducing the computational overhead. The original MAE paper \cite{he2022masked} also emphasizes fine-tuning over full pre-training based on their extensive experiments. Following fine-tuning, we transfer the parameters of the MAE model to a classifier, which is then trained on our limited labeled melt pool image data for the final classification task as shown in Figure \ref{fig2.3}. 
\begin{algorithm}
\caption{Self-superivised Fine-Tuning of MAE}\label{alg1}
\begin{algorithmic}[1]
\State \textbf{Input:} Unlabeled data $\mathcal{X}_\text{unlabeled}$, Pre-trained MAE Encoder $\mathcal{E}_\text{MAE}(\cdot)$ with parameters, $\phi^0$ and Decoder $\mathcal{D}_\text{MAE}(\cdot)$ with parameters,  $\theta^0$
\State \textbf{Output:} Fine-tuned $\phi$ for classifier

\State Initialize MAE with pre-trained $\phi^0$ and $\theta^0$
\State Define the number of epochs for fine-tuning
\For{each epoch}
    \For{each batch of images $x$ in $\mathcal{X}_\text{unlabeled}$}
        \State Patch embedding: ($z,\phi)= \mathcal{E}_{\text{MAE}}(x, \phi^0)$
        \State Reconstruct ($\hat{x}, \theta) = \mathcal{D}_{\text{MAE}}(z, \theta^0)$
        \State Compute Reconstruction Loss (MSE):
        \vspace{-25pt}
        \[
        L_{\text{recon}} = \min_{\phi, \theta} \frac{1}{N} \sum_{i=1}^{N} \| \hat{x}_i - x_i \|^2
        \]\vspace{-25pt}
        \State Update parameters $\phi$ and $\theta$:
        \vspace{-20pt}
        \[
        \phi, \theta \leftarrow \phi, \theta - \eta \nabla_{\phi, \theta} L_{\text{recon}}
        \]\vspace{-30pt}
        \State where, $\eta$ is the learning rate, and $\nabla_{\phi, \theta}$ is the gradient with respect to $\phi$ and $\theta$.
    \EndFor
    \State Save $\phi$ based on the lowest validation loss, $L_{\text{recon}}$
\EndFor
\end{algorithmic}
\end{algorithm}

\subsection{Classifier Models}\label{claasif}
When the fine-tuning is done, we transfer the trained parameters of the MAE Encoder only to our classifiers. We discard the MAE decoder since it is solely the encoder that produces the latent representations for recognition tasks. We transfer the fine-tuned parameters of the MAE Encoder to classifier models to train them in a fully supervised manner to perform the classification task. In this study, we use two different classifiers on the labeled data: a supervised ViT Classifier initialized by the fine-tuned encoder parameters and the fine-tuned MAE Encoder with an MLP classifier trained in a supervised manner, as described below. In this section, we first describe the Vision Transformer (ViT) backbone, which is fundamental in both the MAE Encoder and the ViT Classifier employed in this study. 

\subsubsection{ViT Classifier}\label{vit_class}
The Vision Transformer (ViT) classifier \cite{dosovitskiy2020image} adapts the traditional natural language processing (NLP) Transformer encoder's \cite{vaswani2017attention} structure only, to process images effectively as shown in Figure \ref{fig2.4}. In this approach, each input image of resolution \( H \times W \) with \( C \) color channels is divided into a grid of non-overlapping patches of size \( P \times P \). Each patch is flattened into a 1D vector of size \( P^2 \cdot C \), forming a sequence of patches that serves as the input to the Transformer model. The number of patches is determined by $N = \frac{H \times W}{P^2}$. To clarify the model's functionality, the following section presents and explains key equations from the original ViT paper \cite{dosovitskiy2020image}.

After the image is divided into patches, each patch is then projected into a latent space of fixed dimension $D$ through a trainable linear projection, creating patch embeddings for the Transformer. This process can be represented in equation \ref{eq1} below. 
\begingroup
\small
\begin{equation}\label{eq1} 
z_0 = [x_{\text{class}}; x_p^1 E; x_p^2 E; \dots; x_p^N E] + E_{\text{pos}}, E \in \mathbb{R}^{(P^2 \cdot C) \times D},  E_{\text{pos}} \in \mathbb{R}^{(N+1) \times D}
\end{equation}
\endgroup

\noindent where $x_p$ represents each patch, $E$ is the learnable projection matrix, and $E_{\text{pos}}$ denotes the positional encoding matrix. Learnable 1D positional encodings $E_{\text{pos}}$ are added to each patch embedding to retain spatial information. This allows the Transformer to recognize the spatial relationships between patches. Furthermore, a special classification token $x_{\text{class}}$, similar to the $[CLS]$ token \cite{devlin2018bert} used in natural language processing, is prepended to the sequence. This classification token aggregates information from the entire image and serves as the image's global representation at the output of the Transformer encoder.

The Vision Transformer (ViT) encoder consists of a $L$ stack of Transformer layers, each structured with alternating Multi-Head Self-Attention (MSA) in equation \ref{eq2} and feed-forward Multi-Layer Perceptron (MLP) blocks shown in equation \ref{eq3}. Each block is preceded by layer normalization (LN) and includes residual connections to enhance training stability (Figure \ref{fig2.4}). The classification head consists of an MLP head.  The MSA operation involves computing the attention of multiple heads,  $h$ = \textit{\{ 1, \dots, H \}}. The self-attention mechanism is summarized below in the equation \ref{eq21}.
\vspace{-20pt}
\begin{equation}\label{eq21}
\text{Attention}(Q, K, V) = \text{softmax}\left(\frac{Q K^\top}{\sqrt{d_k}}\right) V
\end{equation}
\vspace{-25pt} 

where, weights \( Q = W^Q [z_i] \), \( K = W^K [z_i] \), \( V = W^V [z_i] \), and \( d_k \) is the dimension of \( K \).

\vspace{3pt} 
For each head \( h \), the attention mechanism is computed as follows:
\vspace{-25pt}
\begin{equation}\label{eq22}
\vspace{-25pt}
\text{head}_h = \text{Attention}\left(z_{(l-1)} W_h^Q, z_{(l-1)} W_h^K, z_{(l-1)} W_h^V\right)
\end{equation}
\vspace{-25pt}
\begin{equation}\label{eq23}
\vspace{-40pt}
\text{MSA}(\cdot) = \text{Concat}(\text{head}_1, \dots, \text{head}_H)
\end{equation}
\vspace{-10pt}
\begin{equation} \label{eq2}
\vspace{-20pt}
z_\ell^0 = \text{MSA}(\text{LN}(z_{\ell-1})) + z_{\ell-1}, \quad \ell = 1, \dots, L 
\end{equation}

In equations \ref{eq21}, \ref{eq22}, \ref{eq23}, the MSA mechanism allows the model to assess relationships between all patches by weighing their interactions respectively, while layer normalization (LN) is applied before MSA to stabilize the learning process (equation \ref{eq2}). The residual connection adds the original input back to the MSA output, aiding in gradient flow and enabling deeper model structures. 
\vspace{-15pt}
\begin{equation} \label{eq3}
\vspace{-15pt}
z_\ell = \text{MLP}(\text{LN}(z_\ell^0)) + z_\ell^0, \quad \ell = 1, \dots, L 
\end{equation}

Here, the MLP block refines each patch's representation individually and consists of two linear layers with a Gaussian Error Linear Unit (GELU) activation in between. The residual connection here also supports stable gradient flow, prevents vanishing gradient, and helps preserve the input features while enhancing the patch representation in equation \ref{eq3}.
\vspace{-20pt}
\begin{equation} \label{eq4}
\vspace{-20pt}
y = \text{LN}(z_L^0) 
\end{equation}

In equation \ref{eq4}, the layer normalization is applied specifically to the class embedding $z_L^0$ from the last layer, which represents the entire image’s complete information. Then, the normalized class token, $y$ serves as the final output representation of the image and is subsequently used for classification by the softmax layer in the MLP head (Figure \ref{fig2.4}). This structure enables the ViT model to capture and integrate both local and global spatial relationships across patches. 

We transfer the fine-tuned weights $\phi$ from the MAE Encoder (Algorithm \ref{alg1}) to the ViT Classifier (Algorithm \ref{alg2}) to initialize the supervised training on the labeled melt pool images. The training steps for the ViT Classifier are outlined in the Algorithm \ref{alg2}. 
\vspace{5pt}

\begin{algorithm}
\caption{Classification with ViT Classifier}\label{alg2}
\begin{algorithmic}[1]
\State \textbf{Input:} Labeled data $\mathcal{X}_\text{labeled}$, labels $y$, Fine-tuned MAE Encoder parameters $\phi$, ViT Classifier encoder $E_\text{ViT}(\cdot)$
\State \textbf{Output:} Class prediction  $\hat{y}$ for each image $x$
\State Divide each image $x$ into patches $\{x_{p}^i\}_{i=1}^N$, where $x_{p}^i \in \mathbb{R}^{P \times P}$
\State Define the number of epochs for training 
\For{each epoch:}
    \State Patch embedding, $z^i = E_\text{ViT}(x_{p}^i, \phi),$ \quad i = 1, \ldots, N
    \State Append class embedding $[CLS]$ from patch embedding sequence for image $x$:
    \vspace{-20pt}
    \[  [CLS, z^1, z^2, \dots, z^N]
    \]\vspace{-30pt}
    \State Obtain class logits from final $[CLS]$:
    \vspace{-20pt}
    \[
      \text{logits} = \, _{\psi^c}{\text{MLP}}([CLS])
    \]\vspace{-30pt}
    \State Apply softmax to obtain class probabilities:
    \vspace{-20pt}
    \[
    \hat{y} = \text{softmax}(\text{logits})= \frac{\exp(\text{logits}_j)}{\sum_{k} \exp(\text{logits}_k)}
    \]\vspace{-25pt}
    \State Compute cross-entropy loss with labels $y$:
    \vspace{-20pt}
    \[
    L_{\text{class}} = - \sum_{c} y_c \log \hat{y}_c
    \]\vspace{-25pt}
    \State Update parameters $\phi$ and $\psi^c$ to minimize $L_{\text{class}}$:
    \vspace{-20pt}
    \[
    \phi, \psi^c \leftarrow \phi, \psi^c - \eta \nabla_{\phi, \psi^c} L_{\text{class}}
    \]\vspace{-30pt}
    \State \textbf{Return} Class predictions $\hat{y}$
\State Save model based on the lowest validation loss, $L_{\text{class}}$
\EndFor
\end{algorithmic}
\end{algorithm}
\vspace{-15pt}
\subsubsection{MAE Encoder classifier}
\begin{figure*}[t]\vspace*{4pt}
    \centering
    \includegraphics[width=0.65\textwidth, height=0.65\textheight]{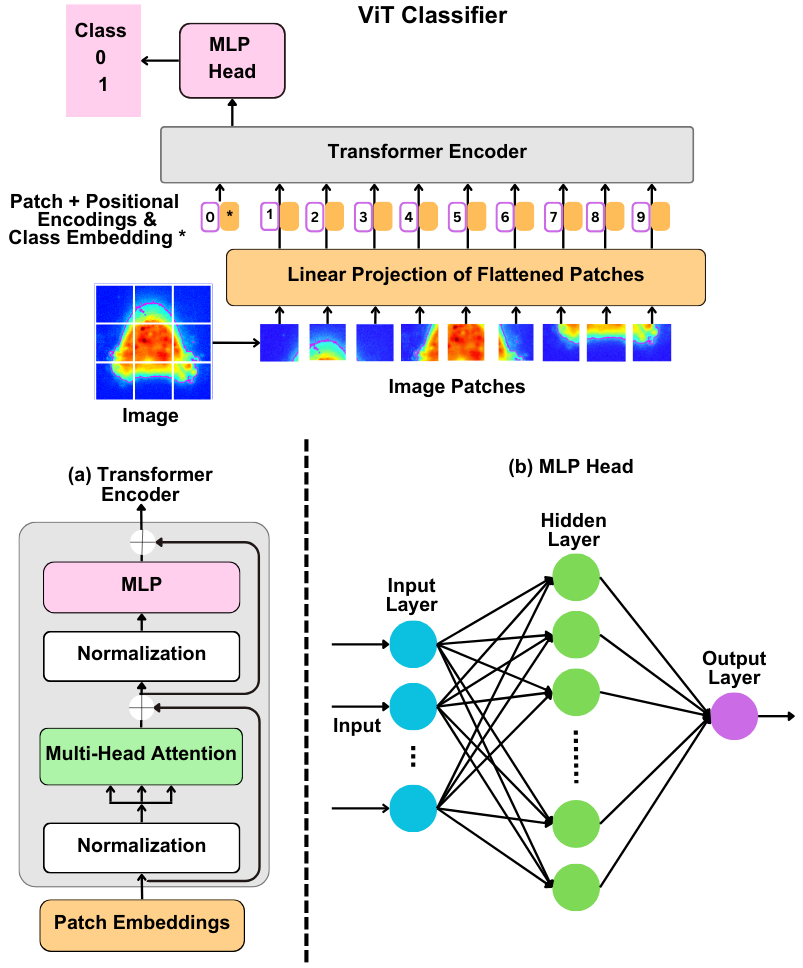}
    \caption{Overview of Vision Transformer Classifier redrawn from \cite{dosovitskiy2020image}; (a) Transformer Encoder Architecture, (b) MLP head Architecture.}
    \label{fig2.4}
\end{figure*}

As we already mentioned earlier, the MAE Encoder has a similar architecture as a ViT encoder (Figure \ref{fig2.4}) except for the MLP classifier head. Thus, we add an extra two-layer MLP classifier head to the end of the MAE Encoder model to perform the classification task here. Additionally, while the ViT classifier incorporates a class embedding and a designated token $[CLS]$ within the input patch embeddings, the MAE Encoder instead utilizes a dummy token during pretraining, which serves a similar function but is primarily intended for alignment with the ViT architecture rather than explicit classification. As advised in the original MAE paper \cite{he2022masked}, applying global pooling over the encoder’s output tokens is recommended to obtain a single image representation for classification. In our implementation, we followed this recommendation and evaluated different pooling strategies. We found that max-pooling \cite{murray2014generalized} of the output features yielded superior performance. Thus, we apply max-pooling to the output embeddings and generate a similar class representation (Algorithm \ref{alg3}). 

We train this setup of fine-tuned MAE Encoder with the MLP classifier extension on the labeled melt pool data. This additional MLP head allows the model to map learned features to the desired class labels, and perform classification. The goal is to evaluate the performance of the fine-tuned MAE Encoder paired with an MLP classifier, leveraging the features learned during the self-supervised fine-tuning phase on unlabeled melt pool images, trained on labeled data to classify melt pool categories effectively. The training steps are demonstrated in the Algorithm \ref{alg3}.

\begin{algorithm}
\caption{Classification with MAE Encoder Classifier}\label{alg3}
\begin{algorithmic}[1]
\State \textbf{Input:} Labeled data $\mathcal{X}_\text{labeled}$, labels $y$, Fine-tuned MAE Encoder $\mathcal{E}_\text{MAE}(\cdot)$ with  parameters $\phi$, 
\State \textbf{Output:} Class prediction $\hat{y}$ for each image $x$
\State Divide each image $x$ into patches $\{x_{p}^i\}_{i=1}^N$, where $x_{p}^i \in \mathbb{R}^{P \times P}$
\State Define the number of epochs for training
\For{each epoch:}
    \State Patch embeddings, $z^i =\mathcal{E}_\text{MAE}(x_{p}^i, \phi), \quad i = 1, \ldots, N$
    \State Apply max pooling to get a single embedding vector:
    \vspace{-20pt}
    \[
    z_{\text{pool}} = \text{MaxPool}([z^1, z^2, \dots, z^N])
    \]\vspace{-30pt}
    \State Pass $z_{\text{pool}}$ to MLP classifier to obtain class logits:
    \vspace{-20pt}
    \[
    \text{logits} = \, _{\psi^m} {\text{MLP}}(z_{\text{pool}})
    \]
    \vspace{-30pt}
    \State Apply softmax to obtain class probabilities:
    \vspace{-20pt}
    \[
    \hat{y} = \text{softmax}(\text{logits}) = \frac{\exp(\text{logits}_j)}{\sum_{k} \exp(\text{logits}_k)}
    \]\vspace{-25pt}
    \State Compute cross-entropy loss with labels $y$:
    \vspace{-20pt}
    \[
    L_{\text{class}} = - \sum_{c} y_c \log \hat{y}_c
    \]\vspace{-25pt}
    \State Update parameters $\phi$ and $\psi^m$ to minimize $L_{\text{class}}$:
    \vspace{-20pt}
    \[
    \phi, \psi^m \leftarrow \phi, \psi^m - \eta \nabla_{\phi, \psi^m} L_{\text{class}}
    \]\vspace{-30pt}
    \State \textbf{Return} Class predictions $\hat{y}$
\State Save model based on the lowest validation loss, $L_{\text{class}}$
\EndFor

\end{algorithmic}
\end{algorithm}

\section{Results \& Discussion}\label{result}
In this section, we discuss the training strategies, experimental setup, models' architectures, and results in detail.  
\subsection{Training setup of self-supervised MAE}\label{training_mae}
\begin{figure*}[t]\vspace*{4pt}
    \centering
    \includegraphics[width=\textwidth]{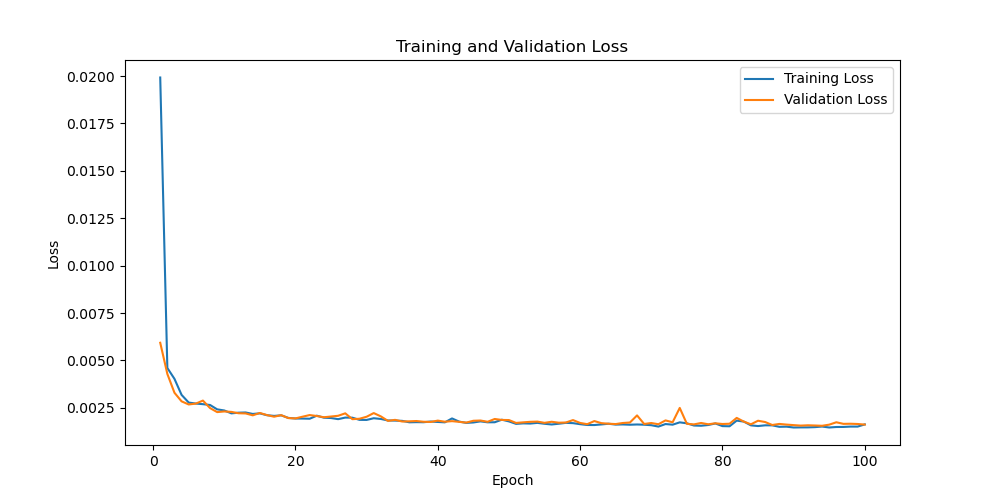}
    \caption{The learning curve of the self-supervised MAE model during fine-tuning on unlabeled melt pool images}
    \label{fig4.1}
\end{figure*}

First, we fine-tune the pre-trained MAE model using 7812 unlabeled melt pool images collected from Sample 1 from Table \ref{tab1} in a self-supervised manner to learn the spatial features through image reconstruction. The data is divided into an 80\%-20\% split for the training and validation.  We use the MAE-base model \cite{mae_github}, configured with an encoder consisting of 12 transformer blocks with a width of 768, and a decoder with 8 Transformer blocks and a width of 512 and reconstruction target without pixel normalization \cite{he2022masked}. The original $400 \times 400$ RGB images are resized to $224 \times 224$ to reduce computational cost and match the dimensions of the pre-trained MAE. To make the model robust, data augmentation techniques such as rotation and flipping are applied. Hyperparameters are selected based on the empirical tuning to ensure stable convergence. A learning rate of $0.0001$ with the Adam optimizer balanced convergence speed and stability, preventing sharp oscillations. Weight decay $(0.00005)$ reduces overfitting by penalizing large weight magnitudes, encouraging smoother representations, while dropout $(0.5)$ enhances regularization by randomly deactivating neurons, preventing reliance on specific features. The training setup and specific hyperparameters are summarized in Table \ref{tab2} and the learning curve obtained is demonstrated in Figure \ref{fig4.1}.

\begin{table}[h!]
\centering
\caption{Training setup and hyperparameters used for the self-supervised fine-tuning of MAE model.}
\begin{tabular*}{\columnwidth}{l @{\extracolsep{\fill}} c}
\hline
\textbf{Name} & \textbf{Value} \\
\hline
Model & MAE-Base \\
Input dimensions & 3 $\times$ 224 $\times$ 224 \\
Patch size & 16 $\times$ 16 \\
Learning rate & 0.0001 \\
Dropout rate & 0.5 \\
Weight decay rate & 0.00005 \\
Optimizer & Adam \\
Epochs & 100 \\
Batch size & 32 \\
Reconstruction loss & Mean Square Loss (MSE) \\
\hline
\end{tabular*}
\label{tab2}
\end{table}

We have used an NVIDIA A10 GPU (Graphic Processing Unit) with 23 GB RAM for this training. The fine-tuning process takes around 25 hours to complete, largely due to the computational demands of reconstructing images. Once the training is completed, we save the parameters of the best-performing model to transfer to the classifiers for the recognition task. This approach aims to boost the classifiers with the enriched features learned from the fine-tuning phase on a similar domain, to map melt pool images to their respective classes accurately. 
\vspace{-7pt}

\subsection{Training setup of supervised Classifiers}\label{training_vit}
At this stage, we employ two different supervised classifiers, i.e., the fine-tuned MAE Encoder only with an MLP classifier head and a supervised ViT Classifier model. The reason behind attempting the MAE Encoder with MLP head is to test the ability of the MAE Encoder for the recognition task which is proven to be efficient on benchmark image datasets \cite{he2022masked}. We also employ a fully supervised ViT Classifier base model \cite{google_vit}. We transfer the fine-tuned parameters of the self-supervised MAE Encoder to both classifiers. The same computational resources outlined in section \ref{training_mae} are used for the training. We run this setup on 1447 labeled melt images collected from Sample 2 from Table \ref{tab1}. The dataset is initially divided into a test set comprising 15\% of the total data, and the remaining 85\% is used for training. 

During the training phase, we employ a six-fold cross-validation approach to ensure robust training and performance evaluation of our models. Within the cross-validation, the data was split into six equal folds, where in each iteration, one fold served as the validation set while the remaining five folds were used for training. This process is repeated six times. We use the same set of hyperparameters in both classifiers' setups and perform end-to-end supervised training on the labeled dataset to perform the binary classification of the normal and abnormal melt pools. The training setup and model specifications are mentioned in the Table \ref{tab3}. 
\begin{table}[h!]
\centering
\caption{Training setup and hyperparameters used for Classifiers.}
\begin{tabular*}{\columnwidth}{l @{\extracolsep{\fill}} c c}
\hline
\textbf{Name} & \textbf{MAE Encoder} & \textbf{ViT Classifier} \\
\hline
Model & MAE-Base & ViT-Base \\
MLP head & Added & Default \\
$[CLS]$ token & Max pooling & Default \\
Input dimensions & 3 $\times$ 224 $\times$ 224 & 3 $\times$ 224 $\times$ 224 \\
Patch size & 16 $\times$ 16 & 16 $\times$ 16 \\
Transformer Blocks & 12 & 12 \\ 
Learning rate & 0.0001 & 0.0001 \\
Dropout rate & 0.5 & 0.5 \\
Weight decay rate & 0.00005 & 0.00005 \\
Optimizer & Adam & Adam \\
Epochs & 100 & 100 \\
Batch size & 16 & 16 \\
Loss & Cross Entropy Loss  & Cross Entropy Loss\\
\hline
\end{tabular*}
\label{tab3}
\end{table}

\subsection{Performance Evaluation}\label{pe}
\begin{figure*}[t]\vspace*{4pt} 
    \centering
    \includegraphics[width=\textwidth]{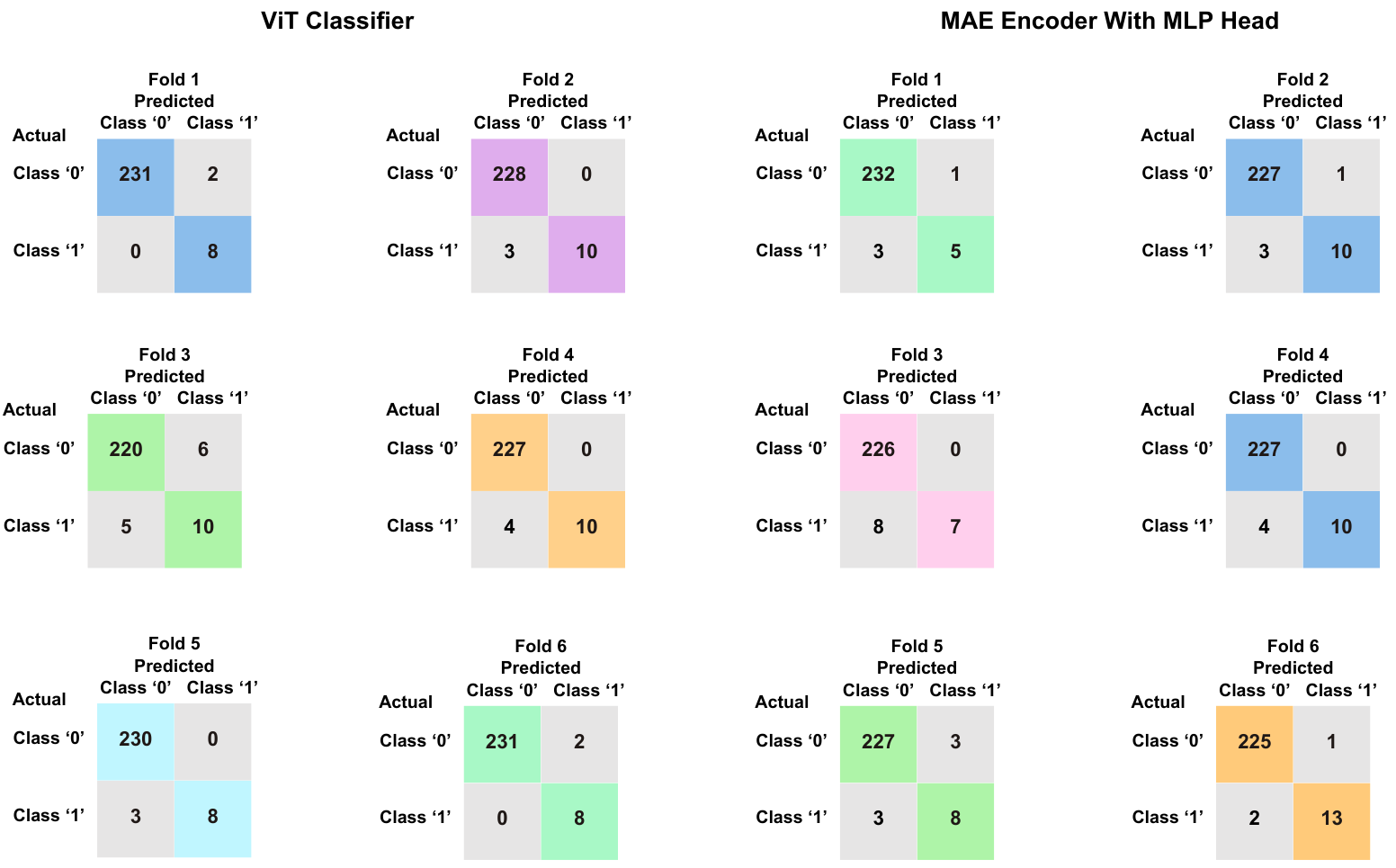}
    \caption{The confusion matrix for each of the folds for ViT Classifier and MAE Encoder Classifier respectively.}\label{fig7}
\end{figure*}

To evaluate the models' performance, we calculate the Accuracy, Precision, Recall, and F1 score for the predicted outcomes. The results from the six-fold cross-validation for both setups are presented in Table \ref{tab4} and Table \ref{tab5}, respectively. Additionally, Figure \ref{fig7} illustrates the confusion matrices across all cross-validation folds.
\vspace{-20pt}
\begin{equation}\label{eq5}
\vspace{-20pt}
\text{Accuracy} = \frac{TP + TN}{TP + TN + FP + FN}
\end{equation}\vspace{-20pt}
\begin{equation}\label{eq6}
\vspace{-20pt}
\text{Precision} = \frac{TP}{TP + FP}
\end{equation}
\vspace{-20pt}
\begin{equation}\label{eq7}
\vspace{-20pt}
\text{Recall} = \frac{TP}{TP + FN}
\end{equation}\vspace{-20pt}
\begin{equation}\label{eq8}
\vspace{-15pt}
\text{F1 Score} = \frac{2 \times \text{Precision} \times \text{Recall}}{\text{Precision} + \text{Recall}}
\end{equation}

Here, \textit{TP = True positive}, the actual positive instances predicted positive; \textit{TN = True Negative}, the actual negative instance predicted negative; \textit{ FP = False Positive}, the actual negative instances predicted positive; and \textit{FN = False Negative}, the actual positive instances predicted negative by the model respectively.  
\begin{table}[h!]
\centering
\caption{Classification results across different folds by ViT classifier.}
\small
\setlength{\tabcolsep}{8pt} 
\begin{tabular}{lcccc}
\hline
\textbf{Fold} & \textbf{Accuracy} & \textbf{Precision} & \textbf{Recall} & \textbf{F1 Score} \\
\hline
1 & 0.9917 & 0.8000 & 1.0000 & 0.8889 \\
2 & 0.9876 & 1.0000 & 0.7692 & 0.8696 \\
3 & 0.9544 & 0.6250 & 0.6667 & 0.6452 \\
4 & 0.9834 & 1.0000 & 0.7143 & 0.8333 \\
5 & 0.9876 & 1.0000 & 0.7273 & 0.8421 \\
6 & 0.9876 & 1.0000 & 0.8000 & 0.8889 \\
\hline
\end{tabular}
\label{tab4}
\end{table}

\begin{table}[h!]
\centering
\caption{Classification results across different folds by MAE Encoder classifier.}
\small
\setlength{\tabcolsep}{8pt}
\begin{tabular}{lcccc}
 \hline
\textbf{Fold} & \textbf{Accuracy} & \textbf{Precision} & \textbf{Recall} & \textbf{F1 Score} \\
\hline
1 & 0.9834 & 0.8333 & 0.6250 & 0.7143 \\
 2 & 0.9854 & 0.9091 & 0.7692 & 0.8333 \\
 3 & 0.9668 & 1.0000 & 0.4667 & 0.6364 \\
4 & 0.9834 & 1.0000 & 0.7143 & 0.8333 \\
 5 & 0.9751 & 0.7273 & 0.7273 & 0.7273 \\
6 & 0.9876 & 0.9286 & 0.8667 & 0.8966 \\
\hline
\end{tabular}
\label{tab5}
\end{table}
\vspace{10pt}

The results from evaluating both classifier models, initiated by the same fine-tuned MAE, over six folds from Figure \ref{fig7}, indicate that both models achieve consistently high accuracy, with scores ranging between 0.9544 and 0.9917. While both classifiers maintain strong performance, the ViT Classifier generally achieves higher precision, often reaching 1.0000, indicating minimal misclassification of negative samples as positives. Recall varies across folds for both models, with each achieving stronger performance in different folds. However, the ViT Classifier demonstrates more consistent recall across folds, ranging from $0.6667 \sim 1.000$, resulting in a slightly higher overall F1 score of $0.6452 \sim 0.8889$. In contrast, the MAE Encoder Classifier shows slightly lower recall scores, ranging from 0.4667 to 0.8667, with F1 scores between $0.7143 \sim 0.8966$. The F1 score reflects the balance between precision and recall, accounting for \textit{FP} and\textit{ FN }. This makes it a better performance indicator than accuracy in cases where one class dominates the test dataset. The F1 scores highlight that the ViT Classifier sometimes produces fewer false positives and negatives, leading to more consistent classification results. Thus, while the MAE Encoder Classifier performs well, the MAE Encoder with the ViT Classifier demonstrates a slight advantage in precision, recall, and F1 score and stability across metrics, making it a preferable choice for melt pool classification.

The slightly lower performance of the MAE Encoder Classifier can be attributed to its default design, which is primarily specialized for spatial feature extraction by reconstruction rather than classification. Since the Vision Transformer (ViT) was originally designed for classification tasks, fine-tuning naturally aligns with its pre-existing architecture, allowing for efficient adaptation. In contrast, the MAE encoder was pre-trained for a reconstruction-based objective, requiring a significant shift in representation learning to accommodate classification. This transition may have resulted in suboptimal adaptation, as the encoder had to reconfigure its learned feature representations for discriminative rather than generative tasks. Consequently, the ViT classifier fine-tuned more effectively, while the MAE encoder required additional adjustments to integrate the newly introduced classification head, potentially leading to slower convergence and slightly lower performance.

\subsection{Limitations \& Future Work}\label{limit}
While our framework achieves high classification accuracy with both models, there are limitations to consider. In this study, we use Mean Squared Error (MSE) as the reconstruction loss; however, other loss functions, such as contrastive loss \cite{wang2021understanding}, could be explored for this data type in future work. The self-supervised MAE model could benefit from an even larger and more diverse set of melt pool images collected under varying process parameters. Although our classifiers performed well, their generalization could improve with more labeled data in both the training and testing phases. 

Furthermore, while ViT models outperform CNNs in capturing complex local and global features, they are computationally intensive, mainly due to quadratic operations like self-attention, leading to longer training times. Class imbalance \cite{johnson2019survey, saini2023tackling} is another challenge: generating high-quality images to balance classes in ViT models requires advanced generative models, which also require significant computational resources. Although class balancing isn't essential for the self-supervised MAE model, it would improve the robustness of the classifiers. If the synthetic data lacks accuracy or quality, the ViT models may struggle to effectively learn features, significantly impacting overall performance. Lastly, to make optimal use of limited labeled data, semi-supervised techniques \cite{van2020survey} can be explored in the future as an alternative to fully supervised classification approaches, potentially enhancing model effectiveness with reduced labeling requirements.

\section{Conclusions}\label{conclu}
The integration of advanced machine vision and AI is transforming additive manufacturing, enabling the handling of complex data types like thermal images of melt pools. These melt pools, which undergo intricate laser-material interactions and rapid solidification, are key indicators of the physical state of a print and the potential for flaw formation. However, their nuanced thermal distributions are challenging to interpret accurately, and labeling them requires extensive ex-situ inspection, which is both time-intensive and costly. To address these challenges, our framework leverages a self-supervised MAE on a large set of unlabeled melt pool images, combined with a supervised classifier trained on limited labeled data. This approach captures both local and global features more effectively than CNNs, making it an ideal fit for melt pool characterization. We summarize the key findings of our study as follows. 
\begin{itemize}
    \item We successfully reduce the dependency on large labeled datasets for supervised training by implementing a self-supervised MAE model to learn the spatial features from the similar but unlabeled melt pool data. By fine-tuning a pre-trained version of this model, we optimized computational resources effectively and yet achieved high accuracy.
    
    \item For a fair comparison, both classifier models used in this study utilize the fine-tuned parameters and learned features respectively, derived from the same self-supervised MAE model. The ViT Classifier demonstrates slightly superior performance over the MAE Encoder Classifier due to its architectural advantage, achieving an average accuracy of 98.2\% and an average F1 score of 82.8\%. 
    
    \item Although tested on melt pool data from the DED process, this framework can be adapted to other thermal imaging data in similar LAM processes, offering a scalable, efficient solution for in-situ melt pool characterization that can complacent against traditional and expensive ex-situ defect characterization methods.
\end{itemize}

Our framework thus provides a robust and efficient alternative for automated melt pool monitoring and characterization in LAM, capable of enhancing quality control with reduced dependency on extensive labeled datasets.

\section*{Acknowledgements}
\small
\textbf{Funding:} This material is based upon work supported by the National Science Foundation under Grant No. 2119654. Any opinions, findings, conclusions, or recommendations expressed in this material are those of the author(s) and do not necessarily reflect the views of the National Science Foundation. 

\textbf{Conflict of interest/Competing interests:} As authors of this work, we declare that we have no conflicts of interest. 

\textbf{Computational Resources:} The authors would like to acknowledge the Pacific Research Platform, NSF Project ACI-1541349, and Larry Smarr (PI, Calit2 at UCSD) for providing the computing infrastructure used in this project.

\textbf{XCT scanning Facility:} The authors express their gratitude to Sarah McLaughlin, Ph.D., and Amanda Stewart, Ph. D. from the AMIF Applications and Imaging Facilities at West Virginia University (WVU) for their assistance in initiating the XCT scanning of the samples. 

\bibliographystyle{elsarticle-num}
\bibliography{ref}

\begin{thebibliography}{10}
\expandafter\ifx\csname url\endcsname\relax
  \def\url#1{\texttt{#1}}\fi
\expandafter\ifx\csname urlprefix\endcsname\relax\def\urlprefix{URL }\fi
\expandafter\ifx\csname href\endcsname\relax
  \def\href#1#2{#2} \def\path#1{#1}\fi

\bibitem{ahn2021directed}
D.-G. Ahn, Directed energy deposition (ded) process: state of the art, International Journal of Precision Engineering and Manufacturing-Green Technology 8~(2) (2021) 703--742.

\bibitem{thompson2015overview}
S.~M. Thompson, L.~Bian, N.~Shamsaei, A.~Yadollahi, An overview of direct laser deposition for additive manufacturing; part i: Transport phenomena, modeling and diagnostics, Additive Manufacturing 8 (2015) 36--62.

\bibitem{svetlizky2021directed}
D.~Svetlizky, M.~Das, B.~Zheng, A.~L. Vyatskikh, S.~Bose, A.~Bandyopadhyay, J.~M. Schoenung, E.~J. Lavernia, N.~Eliaz, Directed energy deposition (ded) additive manufacturing: Physical characteristics, defects, challenges and applications, Materials Today 49 (2021) 271--295.

\bibitem{zhang2024pore}
K.~Zhang, Y.~Chen, S.~Marussi, X.~Fan, M.~Fitzpatrick, S.~Bhagavath, M.~Majkut, B.~Lukic, K.~Jakata, A.~Rack, et~al., Pore evolution mechanisms during directed energy deposition additive manufacturing, Nature Communications 15~(1) (2024) 1715.

\bibitem{tang2020review}
Z.-j. Tang, W.-w. Liu, Y.-w. Wang, K.~M. Saleheen, Z.-c. Liu, S.-t. Peng, Z.~Zhang, H.-c. Zhang, A review on in situ monitoring technology for directed energy deposition of metals, The International Journal of Advanced Manufacturing Technology 108 (2020) 3437--3463.

\bibitem{era2023machine}
I.~Z. Era, M.~A. Farahani, T.~Wuest, Z.~Liu, Machine learning in directed energy deposition (ded) additive manufacturing: A state-of-the-art review, Manufacturing Letters 35 (2023) 689--700.

\bibitem{khanzadeh2018porosity}
M.~Khanzadeh, S.~Chowdhury, M.~Marufuzzaman, M.~A. Tschopp, L.~Bian, Porosity prediction: Supervised-learning of thermal history for direct laser deposition, Journal of manufacturing systems 47 (2018) 69--82.

\bibitem{tian2021deep}
Q.~Tian, S.~Guo, E.~Melder, L.~Bian, W.~â. Guo, Deep learning-based data fusion method for in situ porosity detection in laser-based additive manufacturing, Journal of Manufacturing Science and Engineering 143~(4) (2021) 041011.

\bibitem{khanzadeh2019situ}
M.~Khanzadeh, S.~Chowdhury, M.~A. Tschopp, H.~R. Doude, M.~Marufuzzaman, L.~Bian, In-situ monitoring of melt pool images for porosity prediction in directed energy deposition processes, IISE Transactions 51~(5) (2019) 437--455.

\bibitem{zhao2021automated}
X.~Zhao, A.~Imandoust, M.~Khanzadeh, F.~Imani, L.~Bian, Automated anomaly detection of laser-based additive manufacturing using melt pool sparse representation and unsupervised learning, 'Solid Freeform Fabrication Symposium – An Additive Manufacturing Conference' (2021).

\bibitem{asadi2024process}
R.~Asadi, A.~Queguineur, O.~Wiikinkoski, H.~Mokhtarian, T.~Aihkisalo, A.~Revuelta, I.~F. Ituarte, Process monitoring by deep neural networks in directed energy deposition: Cnn-based detection, segmentation, and statistical analysis of melt pools, Robotics and Computer-Integrated Manufacturing 87 (2024) 102710.

\bibitem{ertay2021process}
D.~S. Ertay, M.~A. Naiel, M.~Vlasea, P.~Fieguth, Process performance evaluation and classification via in-situ melt pool monitoring in directed energy deposition, CIRP Journal of Manufacturing Science and Technology 35 (2021) 298--314.

\bibitem{gaikwad2022multi}
A.~Gaikwad, R.~J. Williams, H.~de~Winton, B.~D. Bevans, Z.~Smoqi, P.~Rao, P.~A. Hooper, Multi phenomena melt pool sensor data fusion for enhanced process monitoring of laser powder bed fusion additive manufacturing, Materials \& Design 221 (2022) 110919.

\bibitem{ouidadi2023real}
H.~Ouidadi, S.~Guo, C.~Zamiela, L.~Bian, Real-time defect detection using online learning for laser metal deposition, Journal of Manufacturing Processes 99 (2023) 898--910.

\bibitem{kong2023development}
J.~H. Kong, S.~W. Lee, Development of melt-pool monitoring system based on degree of irregularity for defect diagnosis of directed energy deposition process, International Journal of Precision Engineering and Manufacturing-Smart Technology 1~(2) (2023) 137--143.

\bibitem{zeng2024classification}
X.~Zeng, S.~Peng, J.~Guo, G.~Chen, J.~Tang, F.~Wang, Classification of melt pool states for defect detection in laser directed energy deposition using fixconvnext model, Measurement Science and Technology 36~(1) (2024) 015201.

\bibitem{chen2023situ}
L.~Chen, X.~Yao, C.~Tan, W.~He, J.~Su, F.~Weng, Y.~Chew, N.~P.~H. Ng, S.~K. Moon, In-situ crack and keyhole pore detection in laser directed energy deposition through acoustic signal and deep learning, Additive Manufacturing 69 (2023) 103547.

\bibitem{bappy2022morphological}
M.~M. Bappy, C.~Liu, L.~Bian, W.~Tian, Morphological dynamics-based anomaly detection towards in situ layer-wise certification for directed energy deposition processes, Journal of Manufacturing Science and Engineering 144~(11) (2022) 111007.

\bibitem{yuan2022method}
J.~Yuan, H.~Liu, W.~Liu, F.~Wang, S.~Peng, A method for melt pool state monitoring in laser-based direct energy deposition based on densenet, Measurement 195 (2022) 111146.

\bibitem{abranovic2024melt}
B.~Abranovic, S.~Sarkar, E.~Chang-Davidson, J.~Beuth, Melt pool level flaw detection in laser hot wire directed energy deposition using a convolutional long short-term memory autoencoder, Additive Manufacturing 79 (2024) 103843.

\bibitem{chen2024situ}
L.~Chen, S.~K. Moon, In-situ defect detection in laser-directed energy deposition with machine learning and multi-sensor fusion, Journal of Mechanical Science and Technology 38~(9) (2024) 4477--4484.

\bibitem{zhang2024machine}
Z.~Zhang, C.~K. Sahu, S.~K. Singh, R.~Rai, Z.~Yang, Y.~Lu, Machine learning based prediction of melt pool morphology in a laser-based powder bed fusion additive manufacturing process, International Journal of Production Research 62~(5) (2024) 1803--1817.

\bibitem{mauricio2023comparing}
J.~Maur{\'\i}cio, I.~Domingues, J.~Bernardino, Comparing vision transformers and convolutional neural networks for image classification: A literature review, Applied Sciences 13~(9) (2023) 5521.

\bibitem{zhuang2020comprehensive}
F.~Zhuang, Z.~Qi, K.~Duan, D.~Xi, Y.~Zhu, H.~Zhu, H.~Xiong, Q.~He, A comprehensive survey on transfer learning, Proceedings of the IEEE 109~(1) (2020) 43--76.

\bibitem{misra2020self}
I.~Misra, L.~v.~d. Maaten, Self-supervised learning of pretext-invariant representations, in: Proceedings of the IEEE/CVF conference on computer vision and pattern recognition, 2020, pp. 6707--6717.

\bibitem{dosovitskiy2020image}
A.~Dosovitskiy, An image is worth 16x16 words: Transformers for image recognition at scale, arXiv preprint arXiv:2010.11929 (2020).

\bibitem{he2022masked}
K.~He, X.~Chen, S.~Xie, Y.~Li, P.~Doll{\'a}r, R.~Girshick, Masked autoencoders are scalable vision learners, in: Proceedings of the IEEE/CVF conference on computer vision and pattern recognition, 2022, pp. 16000--16009.

\bibitem{kirillov2023segment}
A.~Kirillov, E.~Mintun, N.~Ravi, H.~Mao, C.~Rolland, L.~Gustafson, T.~Xiao, S.~Whitehead, A.~C. Berg, W.-Y. Lo, et~al., Segment anything, in: Proceedings of the IEEE/CVF International Conference on Computer Vision, 2023, pp. 4015--4026.

\bibitem{tong2022videomae}
Z.~Tong, Y.~Song, J.~Wang, L.~Wang, Videomae: Masked autoencoders are data-efficient learners for self-supervised video pre-training, Advances in neural information processing systems 35 (2022) 10078--10093.

\bibitem{devlin2018bert}
J.~Devlin, Bert: Pre-training of deep bidirectional transformers for language understanding, arXiv preprint arXiv:1810.04805 (2018).

\bibitem{vaswani2017attention}
A.~Vaswani, Attention is all you need, Advances in Neural Information Processing Systems (2017).

\bibitem{murray2014generalized}
N.~Murray, F.~Perronnin, Generalized max pooling, in: Proceedings of the IEEE conference on computer vision and pattern recognition, 2014, pp. 2473--2480.

\bibitem{mae_github}
F.~Research, \href{https://github.com/facebookresearch/mae}{Mae: Masked autoencoders are scalable vision learners}, accessed: 2023-11-10 (2023).
\newline\urlprefix\url{https://github.com/facebookresearch/mae}

\bibitem{google_vit}
G.~Research, \href{https://github.com/google-research/vision_transformer}{An image is worth 16x16 words: Transformers for image recognition at scale}, accessed: 2023-11-10 (2021).
\newline\urlprefix\url{https://github.com/google-research/vision_transformer}

\bibitem{wang2021understanding}
F.~Wang, H.~Liu, Understanding the behaviour of contrastive loss, in: Proceedings of the IEEE/CVF conference on computer vision and pattern recognition, 2021, pp. 2495--2504.

\bibitem{johnson2019survey}
J.~M. Johnson, T.~M. Khoshgoftaar, Survey on deep learning with class imbalance, Journal of big data 6~(1) (2019) 1--54.

\bibitem{saini2023tackling}
M.~Saini, S.~Susan, Tackling class imbalance in computer vision: a contemporary review, Artificial Intelligence Review 56~(Suppl 1) (2023) 1279--1335.

\bibitem{van2020survey}
J.~E. Van~Engelen, H.~H. Hoos, A survey on semi-supervised learning, Machine learning 109~(2) (2020) 373--440.

\end{thebibliography}

\end{document}